\documentclass[sigconf,natbib=true]{acmart}

\usepackage{caption}
\usepackage{subcaption}
\usepackage{graphicx}
\usepackage{array}
\usepackage{multirow}

\AtBeginDocument{%
  \providecommand\BibTeX{{%
    \normalfont B\kern-0.5em{\scshape i\kern-0.25em b}\kern-0.8em\TeX}}}

\renewcommand\footnotetextcopyrightpermission[1]{} 
\begin{document}

\title{AI Can Be Cognitively Biased: An Exploratory Study on Threshold Priming in LLM-Based Batch Relevance Assessment}

\author{Nuo Chen}
\email{pleviumtan@outlook.com}
\affiliation{%
  \institution{The Hong Kong Polytechnic University}
  \country{HK, China}}

\author{Jiqun Liu}
\email{jiqunliu@ou.edu}
\affiliation{%
  \institution{The University of Oklahoma}
  \country{OK, USA}}

\author{Xiaoyu Dong}
\email{dongxiaoyu7313@gmail.com}
\affiliation{%
  \institution{The Hong Kong Polytechnic University}
  \country{HK, China}}

\author{Qijiong Liu}
\email{liu@qijiong.work}
\affiliation{%
  \institution{The Hong Kong Polytechnic University}
  \country{HK, China}}
  
\author{Tetsuya Sakai}
\email{tetsuyasakai@acm.org}
\affiliation{%
  \institution{Waseda University}
  \city{Tokyo}
  \country{Japan}}

\author{Xiao-Ming Wu}
\email{xiao-ming.wu@polyu.edu.hk}
\affiliation{%
  \institution{The Hong Kong Polytechnic University}
  \country{HK, China}
}

\begin{abstract}
Cognitive biases are systematic deviations in thinking that lead to irrational judgments and problematic decision-making, extensively studied across various fields. Recently, large language models (LLMs) have shown advanced understanding capabilities but may inherit human biases from their training data. While social biases in LLMs have been well-studied, cognitive biases have received less attention, with existing research focusing on specific scenarios. The broader impact of cognitive biases on LLMs in various decision-making contexts remains underexplored. We investigated whether LLMs are influenced by the threshold priming effect in \textit{relevance judgments}, a core task and widely-discussed research topic in the Information Retrieval (IR) coummunity. The priming effect occurs when exposure to certain stimuli unconsciously affects subsequent behavior and decisions. Our experiment employed 10 topics from the TREC 2019 Deep Learning passage track collection, and tested AI judgments under different document relevance scores, batch lengths, and LLM models, including GPT-3.5, GPT-4, LLaMa2-13B and LLaMa2-70B. Results showed that LLMs tend to give lower scores to later documents if earlier ones have high relevance, and vice versa, regardless of the combination and model used. Our finding demonstrates that LLM's judgments, similar to human judgments, are also influenced by threshold priming biases, and suggests that researchers and system engineers should take into account potential human-like cognitive biases in designing, evaluating, and auditing LLMs in IR tasks and beyond. 
\end{abstract}



\maketitle
\section{Introduction}
\begin{figure}[htbp]
    \centering
    \includegraphics[width=.99\linewidth]{
    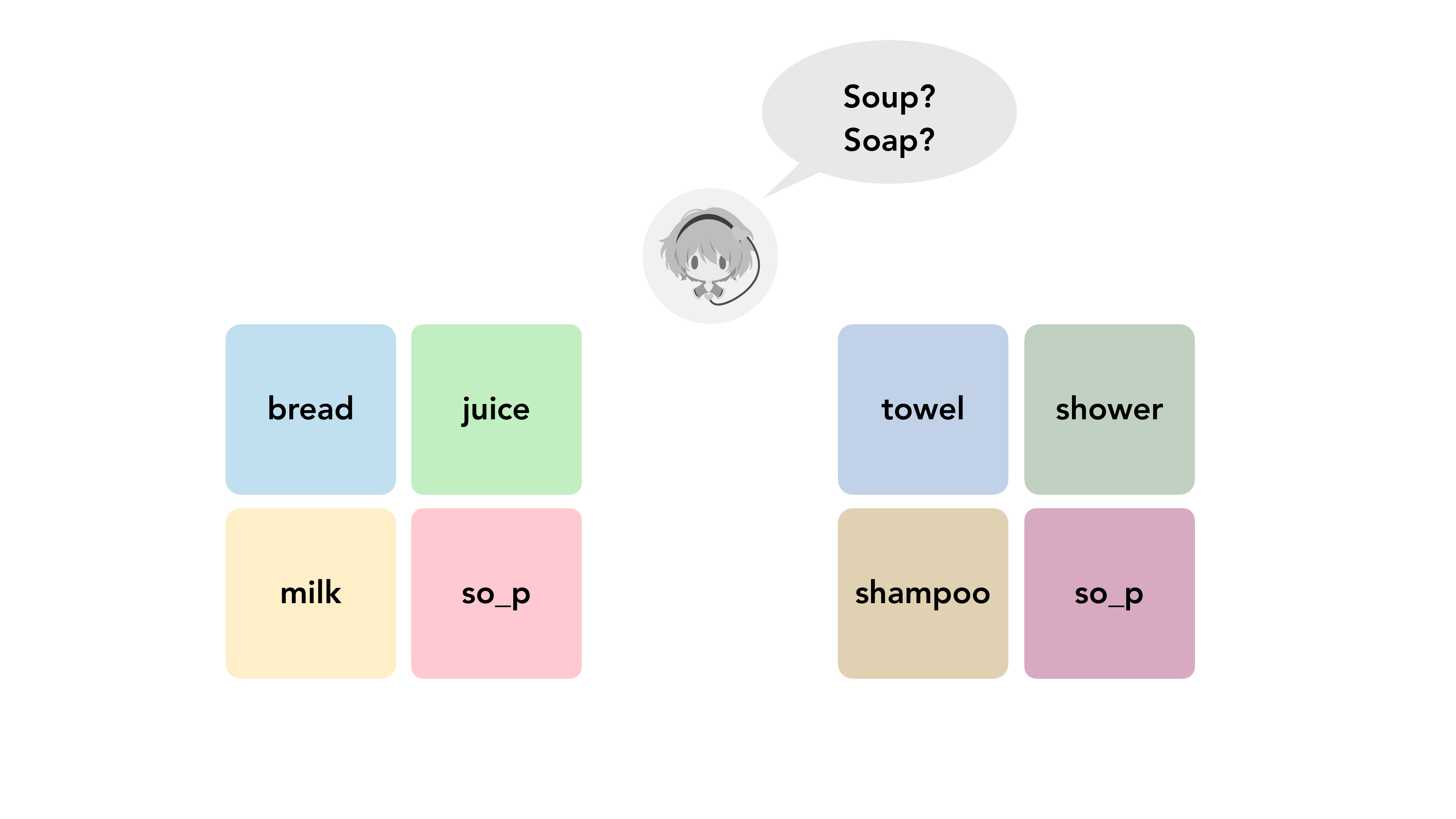
    }
    \caption{\label{fig:example} An example of the priming effect can be illustrated as follows: In the left-hand context, where the stimuli are all related to food, individuals are more likely to associate the stimulus with ``soup.'' Conversely, in the right-hand context, where the stimuli pertain to bathing, individuals are more likely to associate the stimulus with ``soap.''} 
    \Description{
   
    }
\end{figure}

\begin{figure*}[htbp]
    \centering
    \includegraphics[width=.95\linewidth]{
    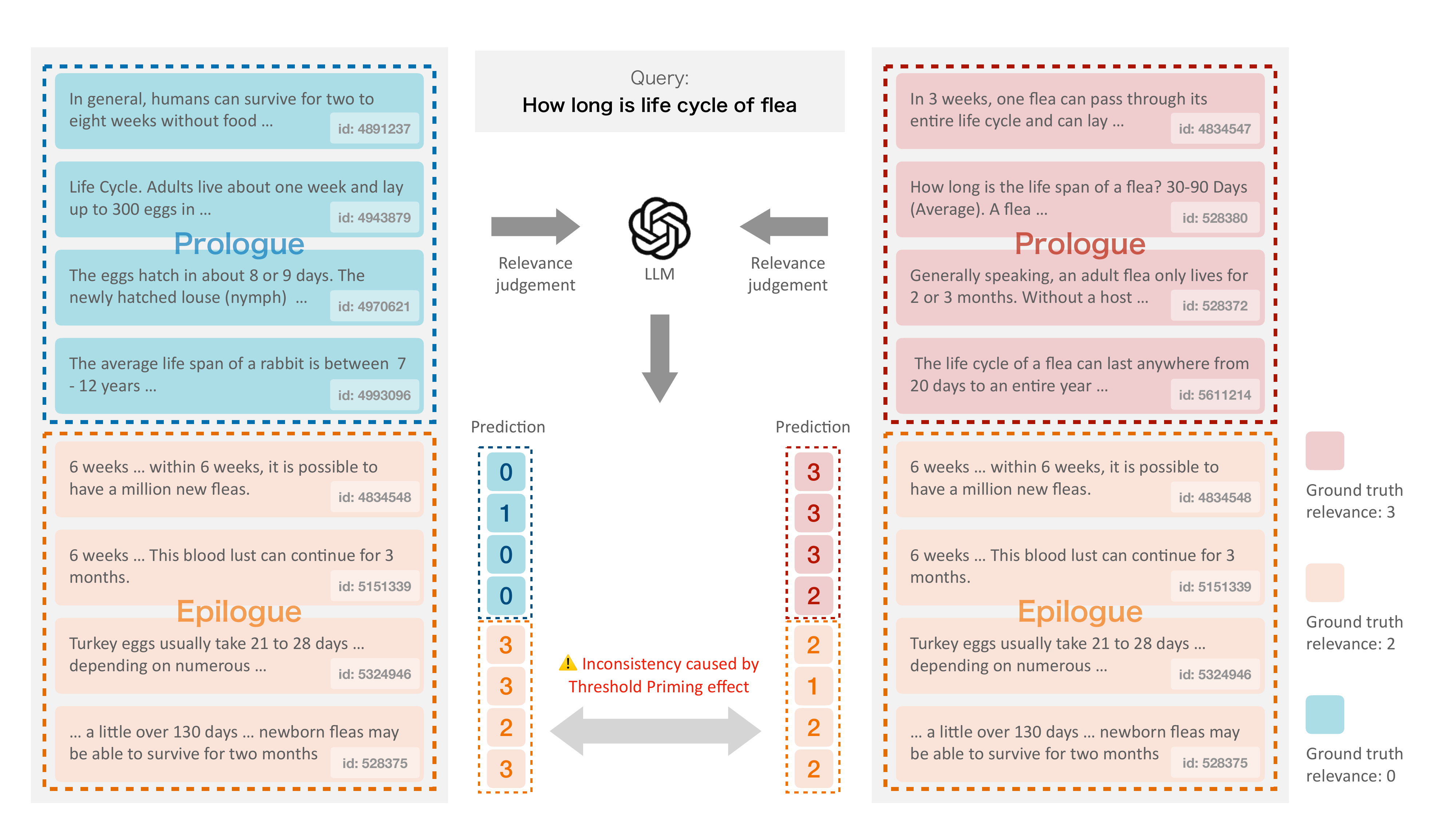
    }
    \caption{\label{fig:method} An example of the methodology adopted in our experiment. In both the left batch and the right batch, the documents comprising the epilogue are identical. However, influenced by the relevance threshold of the documents in the prologue, the LLM assigned different relevance scores to the same documents.} 
    \Description{This illustration shows an experiment investigating the threshold priming effect on LLMs (Large Language Models) when judging the relevance of documents. It includes two sets of documents (prologue and epilogue) containing information about the life cycle and lifespan of fleas. The prologue documents discuss various aspects of flea biology, while the epilogue documents provide additional context. The LLMs are tasked with predicting the relevance of these documents to the query "How long is the life cycle of a flea?" The illustration highlights inconsistencies in predictions caused by the threshold priming effect, showing how the initial prologue documents influence the relevance judgments of the LLMs. The ground truth relevance scores are also provided for comparison.}
\end{figure*}

\textit{\textbf{Cognitive Bias}} refers to a systematic pattern of deviations in thinking 
~\cite{kruglanski1983}, which can lead to irrational judgments and problematic decision-making~\citep{Tversky1974, Tversky1992}. 
This phenomenon has been extensively studied by researchers in fields such as psychology, economics, management, and information science~\citep[e.g.][]{ruggeri2023persistence, enke2023,  Lau-2007, hilbert2012toward, barnes1984}.

In recent years, Large language models (LLMs) have demonstrated advanced natural language understanding capabilities and are increasingly used to support decisions in various tasks~\citep{rastogi2023, li2022pretrained}. 
Nevertheless, having been trained on human-generated data, LLMs may inherit human biases, such as social biases~\citep[e.g.][]{omiye2023large,Zhao2024gender,bang2024measuring,abid2021}, and cognitive biases~\citep[e.g.][]{itzhak2023instructed, enke2023,eicher2024reducing,schmidgall2024addressing}
. Such biases ingrained in their knowledge can be propagated and amplified during human-AI interactions, and consequently, researchers are increasingly concerned with identifying and mitigating biases in LLMs. However, current researches predominantly examine the social biases of LLMs, with cognitive biases receiving comparatively less attention. While some existing work~\citep{echterhoff2024cognitive, schmidgall2024addressing,eicher2024reducing,itzhak2023instructed} has investigated cognitive bias in LLMs during decision-making, these studies have primarily focused on specific types of decisions, such as university admissions~\citep{echterhoff2024cognitive} or medical question answering~\citep{schmidgall2024addressing}, and have only explored a subset of cognitive biases, with the impact of various cognitive biases on LLMs under different types of decision-making scenario remains a critical and promising area awaiting scientific exploration. 

In this study, inspired by previous work~\citep[e.g.][]{scholer2013, Eickhoff2018} in the Information Retrieval (IR) community, we investigate whether LLMs are influenced by the \textbf{priming effect} when making relevance judgments across multiple documents simultaneously. Given that the relevance labels obtained are often used for training IR algorithms and ranking IR systems, relevance labels influenced by cognitive biases could potentially lead to biased algorithms and suboptimal IR systems. Consequently, the IR community has conducted research on how cognitive biases of human assessors, such as the priming effect~\citep{scholer2013}, the anchoring effect~\citep{shokouhi-2015}, the decoy effect~\citep{Eickhoff2018}, affect relevance judgment outcomes. The priming effect is a psychological phenomenon where exposure to certain \textit{stimuli} (such as words, images, or concepts) influences an individual's subsequent behavior, judgments, and decisions unconsciously~\citep{tulving1990priming,VANDERWART198467, herr1986consequences}. Figure~\ref{fig:example} demonstrates an example of the priming effect. In the left-hand context, where the stimuli are all related to food, individuals are more likely to associate the stimulus with ``soup.'' Conversely, in the right-hand context, where the stimuli pertain to bathing, individuals are more likely to associate the stimulus with ``soap.'' Scholer \textit{et al.}~\citep{scholer2013} discovered that when assessors evaluating document relevance, the quality of previously assessed documents can serve as a threshold, triggering the priming effect: if the quality of earlier assessed documents is relatively low, assessors tend to assign higher scores to subsequent documents. As the IR community begins to use LLMs for automatic relevance annotation~\citep{faggioli2023ictir,thomas2023large,upadhyay2024umbrela}, it is necessary to investigate to what extent LLMs are influenced by various contextual factors when making relevance judgments, resulting in biased outcomes that reflect human cognitive biases. Also, more broadly, it is essential to investigate the extent to which LLMs are "rational" in automated judgment tasks, and if they make inaccurate judgments under human cognitive bias triggers. While LLMs could significantly reduce the cost of labeling for Machine Learning tasks both within and beyond IR, we also need to consider the efforts needed for bias mitigation and scalable AI audit.


To address the research gap above, we propose and seek to answer two \textbf{research questions (RQs)} with our experiments:
\begin{itemize}
 \item \textbf{RQ1}: To what extent does the relevance level of earlier documents in a batch influence the relevance judgments of subsequent documents (i.e. threshold priming) when LLMs make batch relevance assessments?
\item \textbf{RQ2}: 
How does the impact of threshold priming in LLMs vary across different topics when making batch relevance judgments of documents?
\end{itemize}

To answer the above RQs, we conducted an experiment on 10 topics from the TREC 2019 Deep Learning (TRDL19) passage retrieval track collection~\citep{craswell2020overview}, with 20 trials for each topic. Figure~\ref{fig:method} illustrates our experimental process. In each trial:~(1) First, we randomly selected $n$ documents with a ground truth relevance of $2$ as the \textit{epilogue};~(2)~Then, we randomly selected $m$ documents with ground truth relevance scores of 3 and 0 to serve as the \textit{high threshold (HT)} and \textit{low threshold (LT) prologues}, respectively;~(3)~We combined the high threshold prologue and low threshold prologue with the same epilogue separately, and then had the LLM judge the relevance scores of each document in these two batches
. 

In our experiment, we tried different combinations of prologue and epilogue lengths, such as prologue length (PL) of 4 and epilogue length (EL) of 4, PL of 4 and EL of 8, and PL of 8 and EL of 8; as well as different widely applied and assessed models, including GPT-3.5, GPT-4o, LLaMa-13B and LLaMa-70B. The results show that:~(1)~When PL is 4 and EL is 4, among most topics, all models exhibited threshold priming in relevance judgment, similar to the observations made by Scholer~\textit{et al.}~\citep{scholer2013}. Specifically, if the ground truth relevance of the prologue is high, the LLMs tend to give lower scores to the documents in the epilogue, and \textit{vice versa}. (2) When PL is 4 and EL is 8, for most topics, LLaMa2-70B did not show significant differences in relevance judgments between the group of HT and the group of LT. However, other models still exhibited threshold priming, tending to give higher relevance scores to the LT group, although LLaMa2-13B is less affected compared to GPT series. (3) When PL is 8 and EL is 8, all models except for LLaMa2-70B tended to give higher relevance scores to the LT group, but LLaMa2-70B tended to give higher relevance scores to the HT group, which is contrary to the trend observed by Scholer \textit{et al.}~\citep{scholer2013}. 

In summary, our primary contributions include:
\begin{itemize}
    \item Our results demonstrate that LLMs can be influenced by threshold priming effect, producing systematically biased relevance assessments. This extends the AI community's understanding of hidden cognitive biases in LLMs, which could be inherited from human-generated data, and encourages further research into their decision-making strategies and thresholds in varying evaluation scenarios.
    \item Our research explores the threshold priming effect in relevance judgment tasks using a variety of LLMs and document batch combinations. Our findings can inspire the IR community to develop methods to identify and mitigate cognitive biases in LLMs in information evaluation tasks.
\end{itemize}

\section{Related Work}
Research in cognitive psychology and behavioral economics indicates that \textit{cognitive biases} emerge from the limitations of human cognitive capacity, particularly when there are insufficient resources to adequately gather and process available information~\citep{kruglanski1983}. Cognitive biases can cause an individual's decisions in uncertain situations to systematically deviate from the expectations of rational decision-making models~\citep{Tversky1974, Tversky1991, Tversky1992}. Typical cognitive biases include the anchoring effect~\citep{Tversky1974}, the decoy effect~\citep{Huber1982}, the reference dependent effect~\citep{kahneman1991anomalies}, the confirmation bias~\citep{SCHWIND20122280}, and the priming effect~\citep{VANDERWART198467}. In this study, we shed light on the \textbf{priming effect}, which is a psychological phenomenon where exposure to certain stimuli (such as words, images, or concepts) influences an individual's subsequent behavior, judgments, and decisions unconsciously~\citep{tulving1990priming,VANDERWART198467, herr1986consequences}. 

\subsection{Cognitive Biases in the LLMs}
Although LLMs do not possess human physiological structures, they may learn cognitive biases expressed in textual form from data and human feedback during the training or fine-tuning process~\citep{itzhak2023instructed}. Current research predominantly examines the social biases of LLMs, with cognitive biases receiving comparatively less attention. There has been work studying the anchoring effect~\citep{echterhoff2024cognitive}, the confirmation bias~\citep{schmidgall2024addressing}, the selection bias~\citep{eicher2024reducing}, and the decoy effect~\citep{itzhak2023instructed} in LLMs, but there are still many types of cognitive biases awaiting exploration, such as the priming effect. Furthermore, existing research on cognitive biases in LLMs is limited to specific scenarios, such as university admission decisions~\citep{echterhoff2024cognitive} or medical question answering~\citep{schmidgall2024addressing}, and how cognitive biases affect LLMs' decision-making in other scenarios remains unknown. In this study, we investigate whether LLMs are influenced by the priming effect when making search decisions such as relevance judgement in place of humans. 

\subsection{Cognitive Biases in Human Search Process}
Humans need to seek information to support various decisions, ranging from everyday decisions such as choosing a good Chinese restaurant in Flushing to major decisions such as deciding whether to undergo a surgery to remove a tumor. In the process of information seeking and interacting with information retrieval systems, individual users also often engage in various decision-making processes, such as, whether to continue browsing the Search Engine Result Page (SERP) or to end the current search~\citep{wicaksono2018,zhang2018requery,chen2021reformulation, liu2017scroll}, determining whether a document is useful in meeting their information needs~\citep{mao2016usefulness, mao2017understanding}, and evaluating the interaction experience with an search system~\citep{jiang2015,siro2023,liu2018satisfaction}. Therefore, cognitive biases have been studied in a variety of works in the field of Information Retrieval (IR)~\citep{Azzopardi2021, liu2023behavioral}. For instance, how the recency effect and the reference-dependent effect influence users' perceptions of search satisfaction~\citep{liu2020reference, Liu-2019-KDD}; and how the anchoring effect, decoy effect, expectation confirmation, and priming effect influence users' judgments of document relevance~\citep{shokouhi-2015, scholer2013, Eickhoff2018, wang2023investigating, wang2024understanding}. Researchers have also explored the potential impact of cognitive biases on socio-political view~\citep{knobloch2015,Kulshrestha2017}, learning and interest development~\cite{liu2021interest}, and personal health decisions~\citep{white2013caption,zhang2012consumer} during the information-seeking process. Additionally, they have investigated how to adjust oversimplified rational user models based on cognitive biases to design evaluation metrics that better characterize user judgments and interaction experience~\citep{zhang2020cascade,chen2022constructing,chen2023reference,chen2024decoy, liu2019investigating}. Among these studies, the work most closely related to our study is probably Scholer \textit{et al.}~\cite{scholer2013}, arguing that assessors, when shown examples of low relevance items, tended to rate subsequent items more highly than if they were shown highly or moderately relevant items first. 

Recently, with Large Language Models~(LLMs) such as GPT-4~\footnote{https://openai.com/index/gpt-4/} and Llama-2~\citep{meta2023llama} demonstrating advanced text comprehension capabilities across various tasks such as text annotation~\citep{Gilardi_2023}, researchers have begun exploring the automation of relevance annotation process using LLMs~\citep{faggioli2023ictir,thomas2023large,upadhyay2024umbrela}. Given that the relevance labels obtained are often used for training IR algorithms and ranking IR systems, relevance labels influenced by cognitive biases could potentially lead to biased algorithms and suboptimal IR systems. However, at this stage, the IR community has a very limited understanding of how cognitive biases in LLMs affect their relevance assessment. In this study, we reveal that LLMs are also influenced by threshold priming, leading to inconsistencies in batch evaluation results (as shown in Figure~\ref{fig:method}).

\section{Research Questions}
To investigate to what extent are LLMs influenced by threshold priming when making batch relevance judgments of documents, in this study, we propose the following \textbf{research questions (RQs)}:
\begin{itemize}
 \item \textbf{RQ1}: To what extent does the relevance level of earlier documents in a batch influence the relevance judgments of subsequent documents (i.e. threshold priming) when LLMs make batch relevance assessments for IR evaluation?
\item \textbf{RQ2}: 
How does the impact of threshold priming in LLMs vary across different topics when making batch relevance judgments of documents?
\end{itemize}

The following sections will introduce our methodology (e.g. experimental setup) and results in response to the RQs above.

\section{Experimental Setup}
\begin{table*}[htbp]
    \centering
    \begin{tabular}{cccccc}
        \hline
        \textbf{topic\_id} & \textbf{query} & \textbf{\#rel=0} & \textbf{\#rel=1} & \textbf{\#rel=2} & \textbf{\#rel=3} \\
        \hline
        168216 & does legionella pneumophila cause pneumonia & 293 & 89 & 128 & 72 \\
        183378 & exons definition biology & 222 & 54 & 15 & 160 \\
        264014 & how long is life cycle of flea & 171 & 59 & 130 & 22 \\
        443396 & lps laws definition & 94 & 31 & 48 & 15 \\
        451602 & medicare's definition of mechanical ventilation & 66 & 54 & 35 & 65 \\
        833860 & what is the most popular food in switzerland & 82 & 33 & 25 & 17 \\
        915593 & what types of food can you cook sous vide & 100 & 13 & 49 & 30 \\
        1112341 & what is the daily life of thai people & 81 & 23 & 22 & 97 \\
        1114819 & what is durable medical equipment consist of & 129 & 128 & 194 & 19 \\
        1117099 & what is a active margin & 138 & 36 & 38 & 45 \\
        \hline
    \end{tabular}
    \caption{Filtered 10 TRDL19 topics: ID, query text, and document counts by relevance levels 0, 1, 2, and 3}
    \label{tab:topic_detail}
\end{table*}

\begin{figure*}[htbp]
    \centering
    \includegraphics[width=.99\linewidth]{
    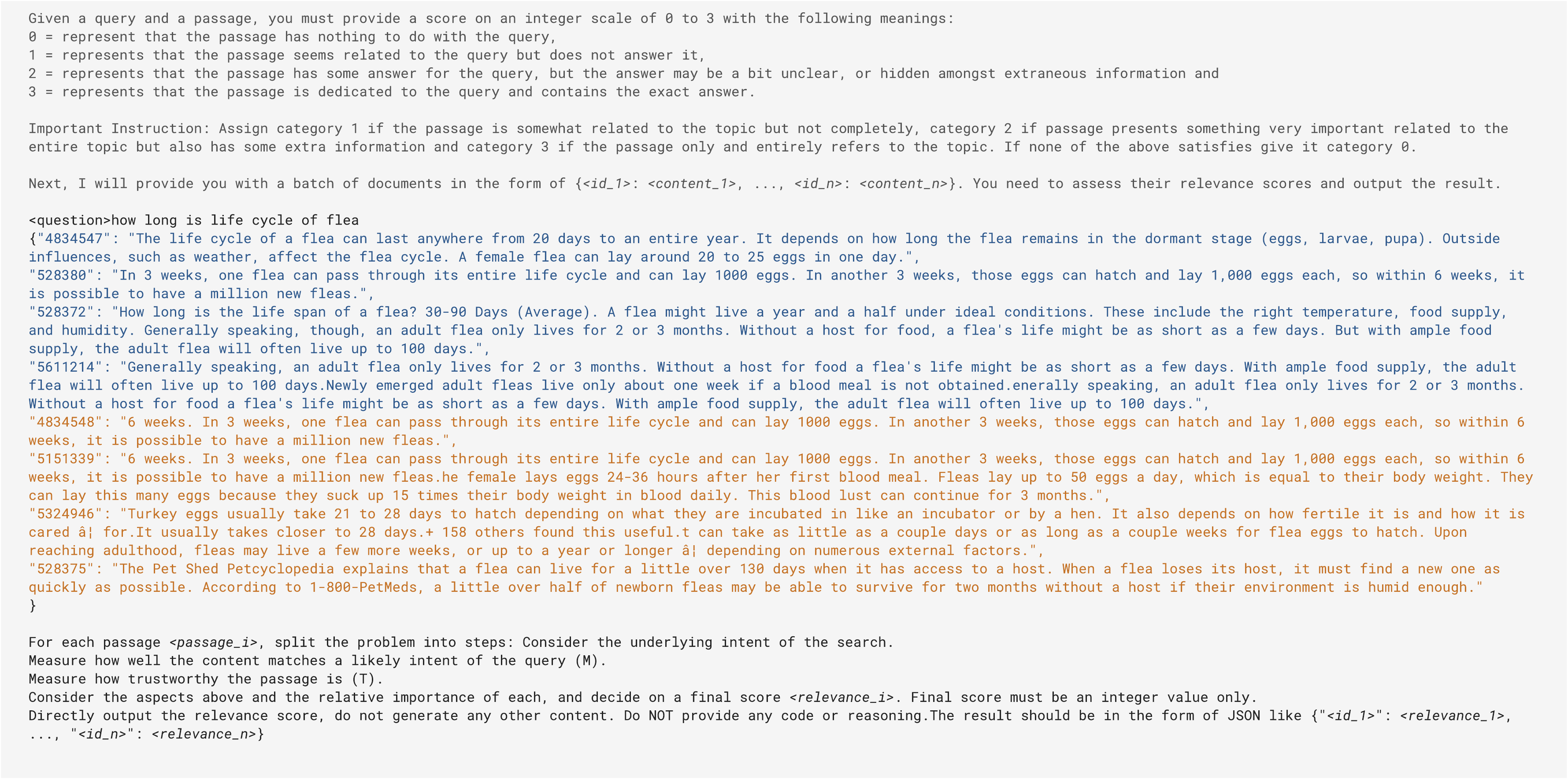
    }
    \caption{\label{fig:prompt} An example of the prompt used in our experiment. 
This is a low threshold batch. The gray text is system prompt, the blue text is the prologue composed of documents with a ground truth relevance of 0, and the dark orange text is the epilogue.} 
    \Description{
    This document provides instructions for assessing the relevance of passages to a given query using a scale of 0 to 3. It explains the meaning of each score, ranging from completely unrelated (0) to perfectly relevant (3). The instructions emphasize considering the passage's match to the query's intent and its trustworthiness. An example query with multiple document passages is provided, and assessors are instructed to output their relevance scores in a JSON format.
    }
\end{figure*}

In this section, we briefly introduce our methodology and experimental setup.  

We conduct our experiment on the TREC 2019 Deep Learning (TRDL19) passage retrieval track collection~\citep{craswell2020overview}. For the sake of convenience, we will refer to a passage as a ``document'' in the following sections. To ensure the selection of topics with a sufficient number of documents at various levels of relevance scores, we filter the topics based on the criterion that there are at least 12 documents for each relevance label (0, 1, 2, and 3). This filtering process results in a final selection of ten topics, which covers a diverse set of domains. In Table~\ref{tab:topic_detail}, the topics are listed with their corresponding topic IDs, query texts, and the number of documents rated as relevance levels 0, 1, 2, and 3 for each topic.

For each topic, we conduct 20 trials. For each trial, we first randomly select $n$ documents with a relevance score of 2, to form the epilogue $\mathbf{E}$:

$$
\mathbf{E} = \{d^e_{1}, d^e_{2}, \ldots, d^e_{n}\}.
$$
Each $d^e_i$ is a document with a ground truth relevance score of 2.

Next, we randomly select $m$ documents with a relevance score of 0 and another $m$ documents with a relevance score of 3 to form the low threshold (LT) prologue $\mathbf{L}$ and the high threshold (HT) prologue $\mathbf{H}$, respectively:

$$
\mathbf{L} = \{d^l_{1}, d^l_{2}, \ldots, d^l_{m}\},
$$
$$
\mathbf{H} = \{d^h_{1}, d^h_{2}, \ldots, d^h_{m}\}.
$$
Each $d^l_i$ and is $d^h_i$ is a document with a ground truth relevance score of 0 and 3 respectively.

We then concatenate $\mathbf{L}$ with $\mathbf{E}$ and $\mathbf{H}$ with $\mathbf{E}$ to obtain the low threshold batch $\mathbf{B_l}$ and the high threshold batch $\mathbf{B_h}$, respectively:
$$
\mathbf{B_l} = \mathrm{concate}\left(\mathbf{L}, \mathbf{E}\right),
$$
$$
\mathbf{B_h} = \mathrm{concate}\left(\mathbf{H}, \mathbf{E}\right).
$$

Finally, for $\mathbf{B_l}$ and $\mathbf{B_h}$, we use the prompts shown in Figure~\ref{fig:prompt} to have a LLM assess the relevance of each document in the batch, and output the results in JSON format. Our prompt design is inspired by and adapted from the work of Upadhyay~\textit{et al.}~\citep{upadhyay2024umbrela}. 

In our experiment, we try different combinations of prologue and epilogue lengths, such as prologue length (PL) of 4 and epilogue length (EL) of 4, PL of 4 and EL of 8, and PL of 8 and EL of 8; as well as different models like GPT-3.5, GPT-4o, LLaMa-13B and LLaMa-70B. We utilized the \texttt{gpt-3.5-turbo}~\footnote{https://platform.openai.com/docs/models/gpt-3-5-turbo} and \texttt{gpt-4}~\footnote{https://platform.openai.com/docs/models/gpt-4o} APIs provided by OpenAI, as well as the \texttt{llama-2-13b-chat}~\footnote{https://replicate.com/meta/llama-2-13b-chat/api} and the \texttt{llama-2-70b-chat}~\footnote{https://replicate.com/meta/llama-2-70b-chat/api} APIs offered by Replicate. For GPT-3.5 and GPT-4, we set the \texttt{temperature} to 0 and \texttt{top\_p} to 1, following the work of Upadhyay \textit{et al.}~\citep{upadhyay2024umbrela}; For LLaMa-2-13B and LLaMa-2-70B, we set the \texttt{temperature} to 0, \texttt{top\_p} to 1, \texttt{frequency\_penalty} to 1, and \texttt{presence\_penalty} to 0. 

\section{Experimental Result}
\begin{figure*}[htbp]
\centering
\begin{subfigure}[t]{0.24\textwidth}
    \centering
    \includegraphics[width=4.1cm]{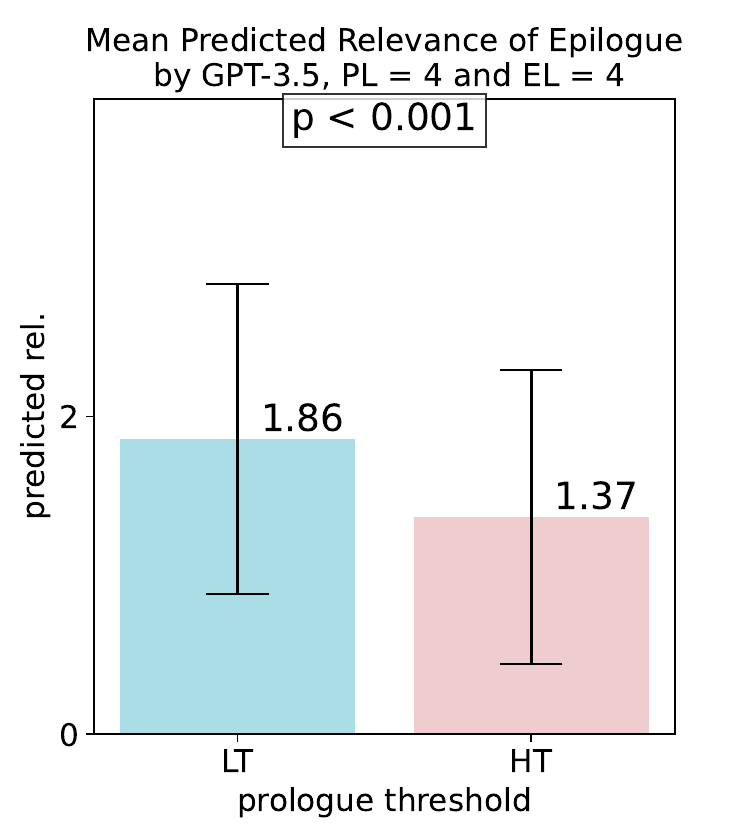}
    \caption*{}
\end{subfigure}
\begin{subfigure}[t]{0.24\textwidth}
    \centering
    \includegraphics[width=4.1cm]{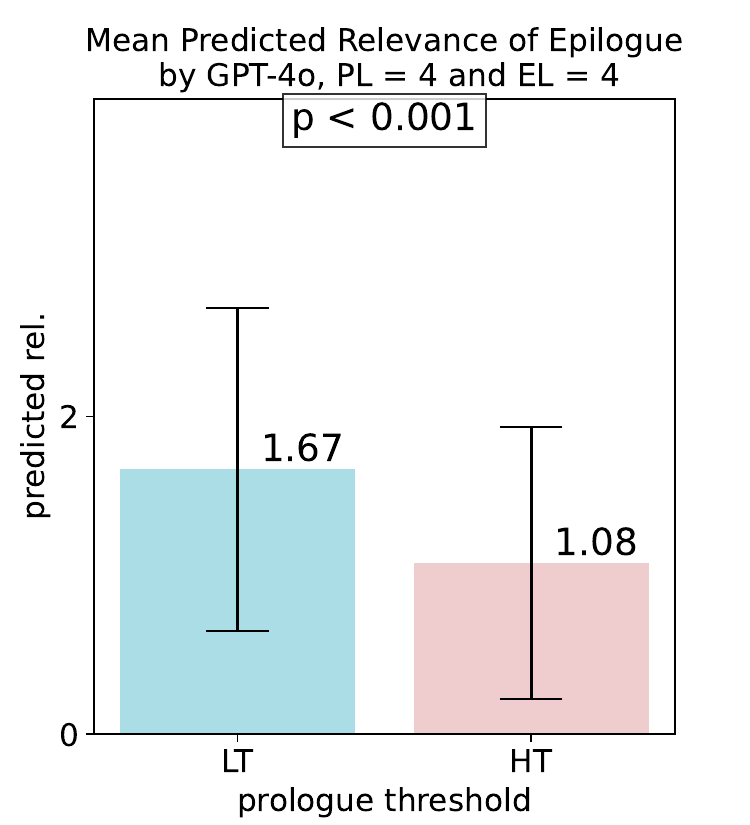}
    \caption*{}
\end{subfigure}
\begin{subfigure}[t]{0.24\textwidth}
    \centering
    \includegraphics[width=4.1cm]{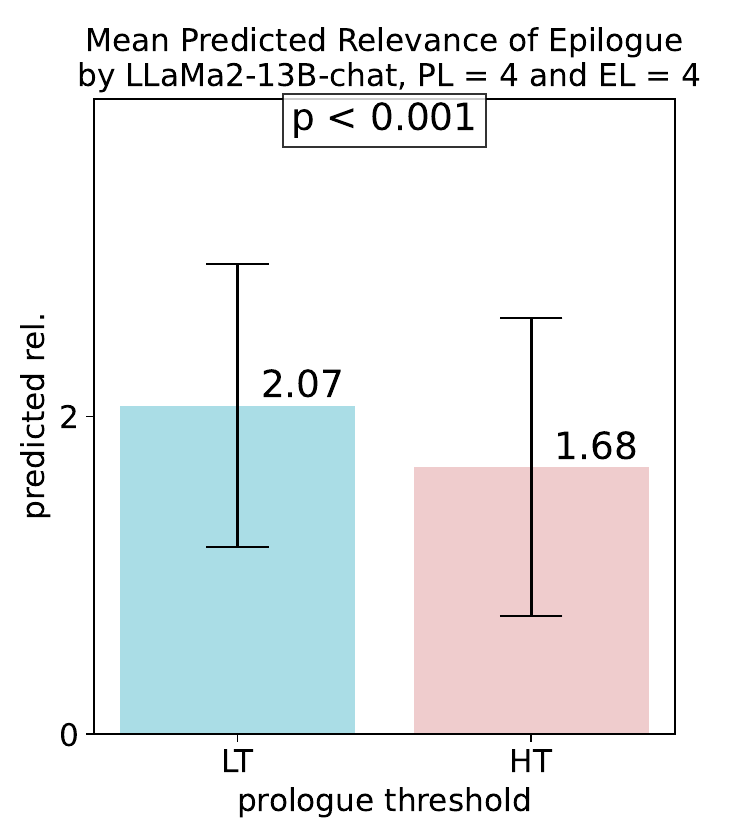}
    \caption*{}
\end{subfigure}
\begin{subfigure}[t]{0.24\textwidth}
    \centering
    \includegraphics[width=4.1cm]{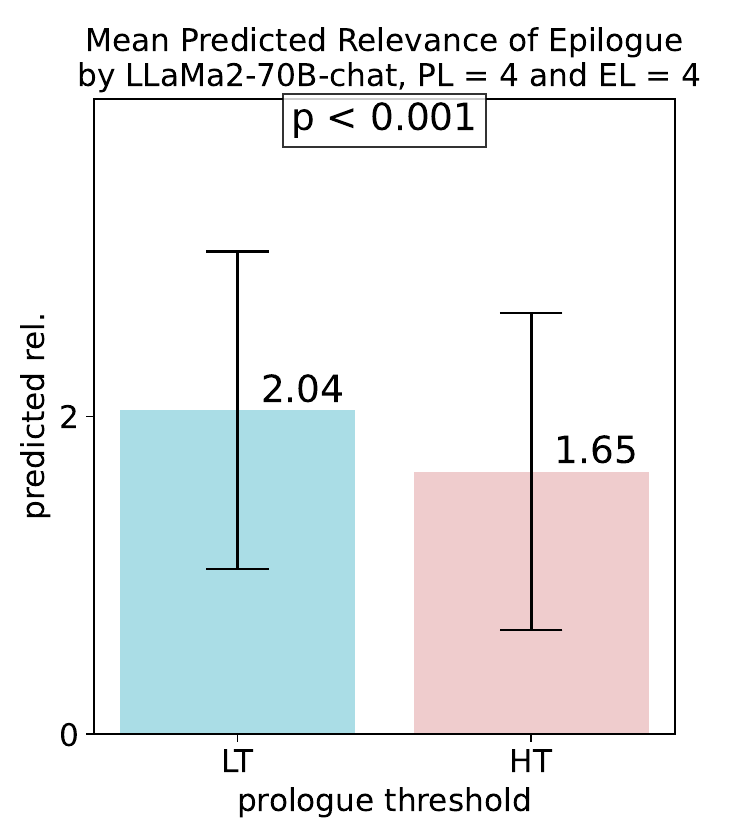}
    \caption*{}
\end{subfigure}
\begin{subfigure}[t]{0.24\textwidth}
    \centering
    \includegraphics[width=4.1cm]{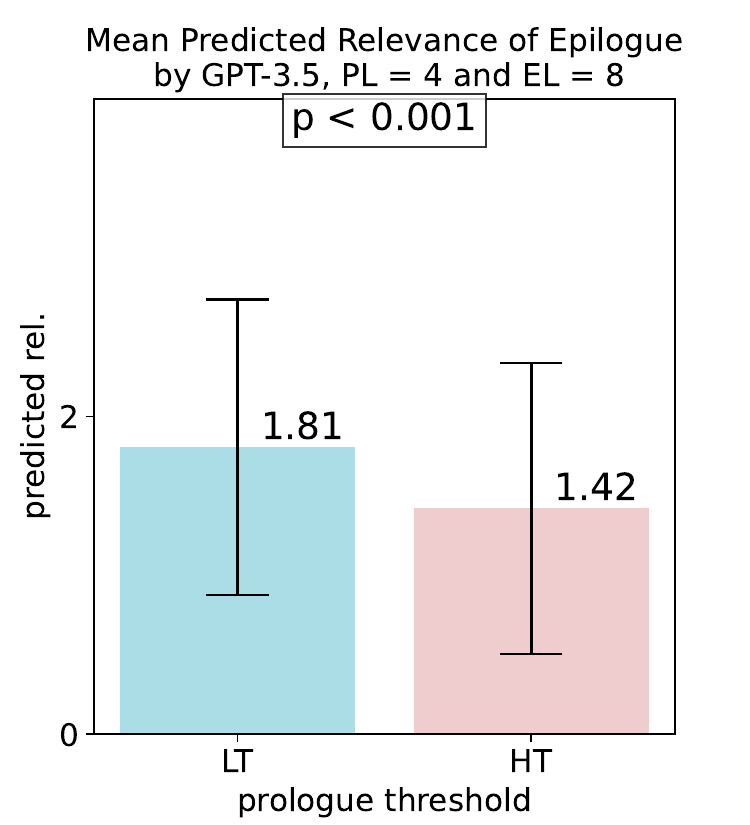}
    \caption*{}
\end{subfigure}
\begin{subfigure}[t]{0.24\textwidth}
    \centering
    \includegraphics[width=4.1cm]{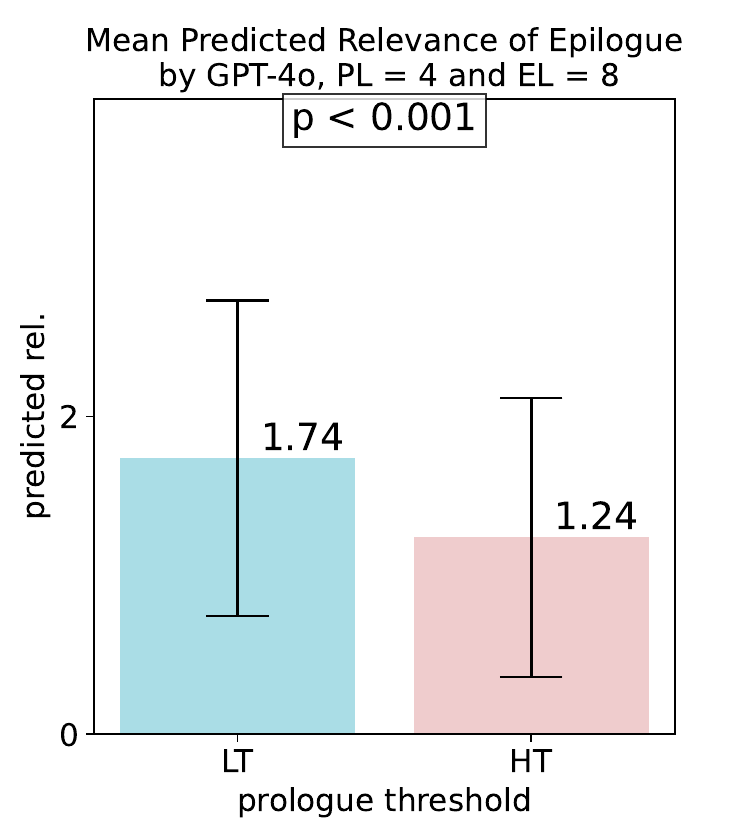}
    \caption*{}
\end{subfigure}
\begin{subfigure}[t]{0.24\textwidth}
    \centering
    \includegraphics[width=4.1cm]{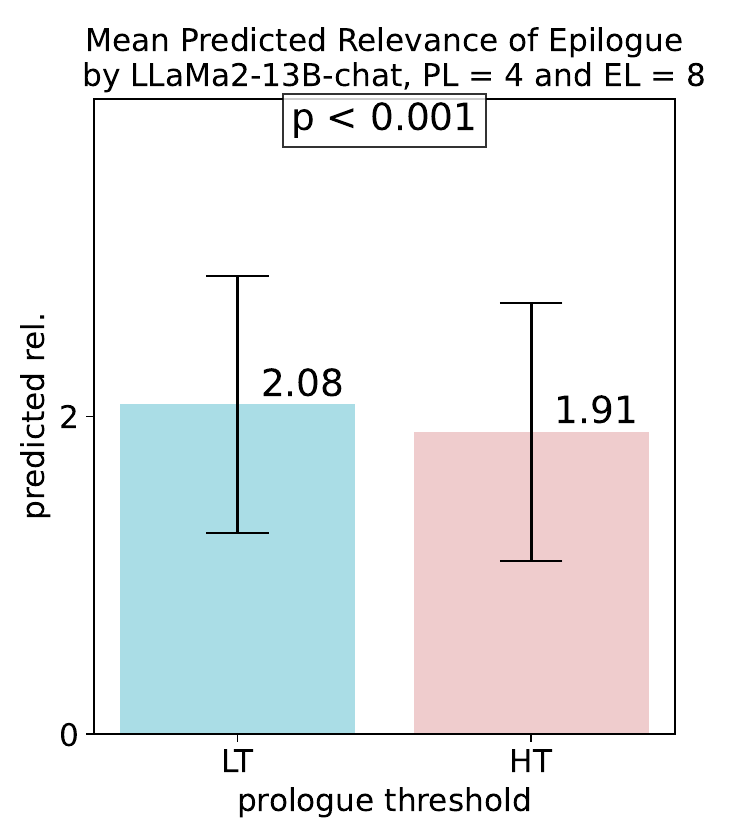}
    \caption*{}
\end{subfigure}
\begin{subfigure}[t]{0.24\textwidth}
    \centering
    \includegraphics[width=4.1cm]{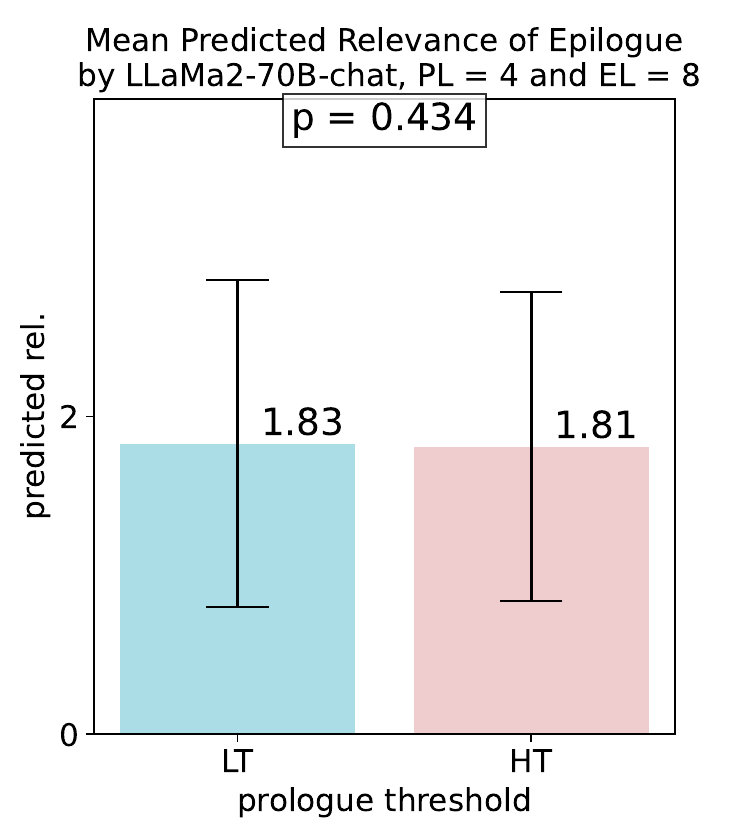}
    \caption*{}
\end{subfigure}
\begin{subfigure}[t]{0.24\textwidth}
    \centering
    \includegraphics[width=4.1cm]{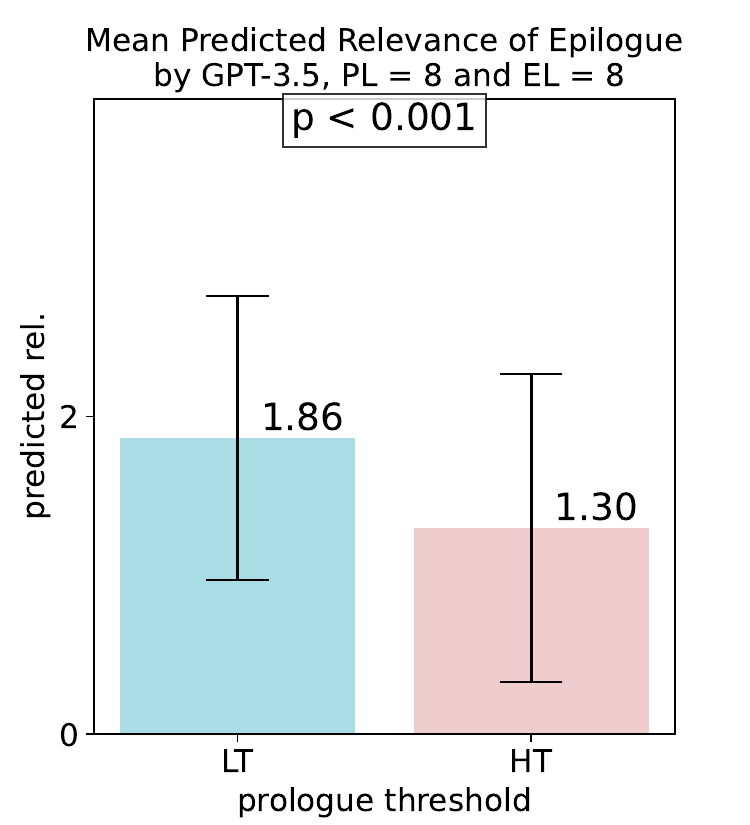}
    \caption*{}
\end{subfigure}
\begin{subfigure}[t]{0.24\textwidth}
    \centering
    \includegraphics[width=4.1cm]{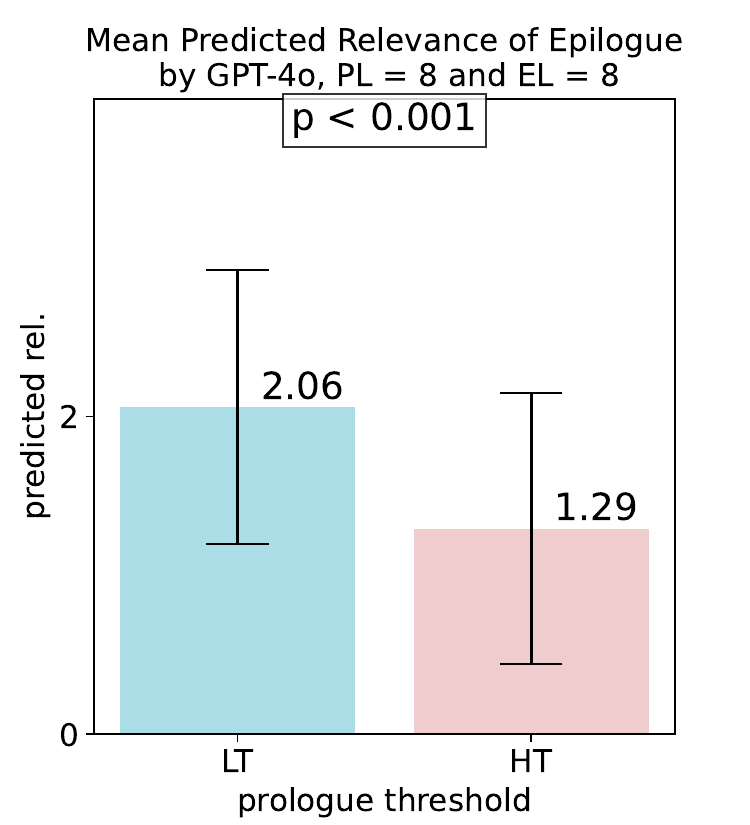}
    \caption*{}
\end{subfigure}
\begin{subfigure}[t]{0.24\textwidth}
    \centering
    \includegraphics[width=4.1cm]{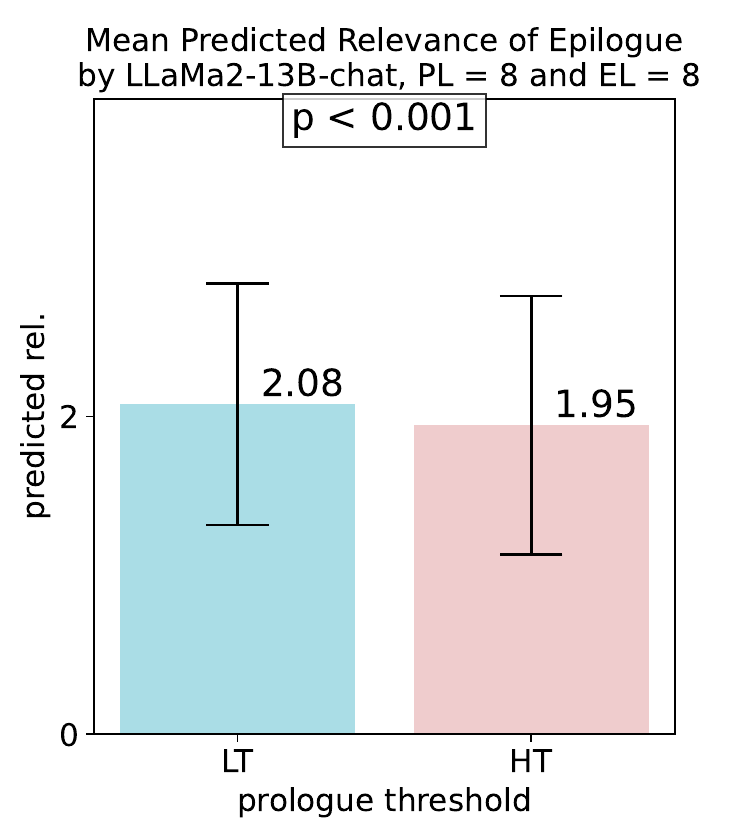}
    \caption*{}
\end{subfigure}
\begin{subfigure}[t]{0.24\textwidth}
    \centering
    \includegraphics[width=4.1cm]{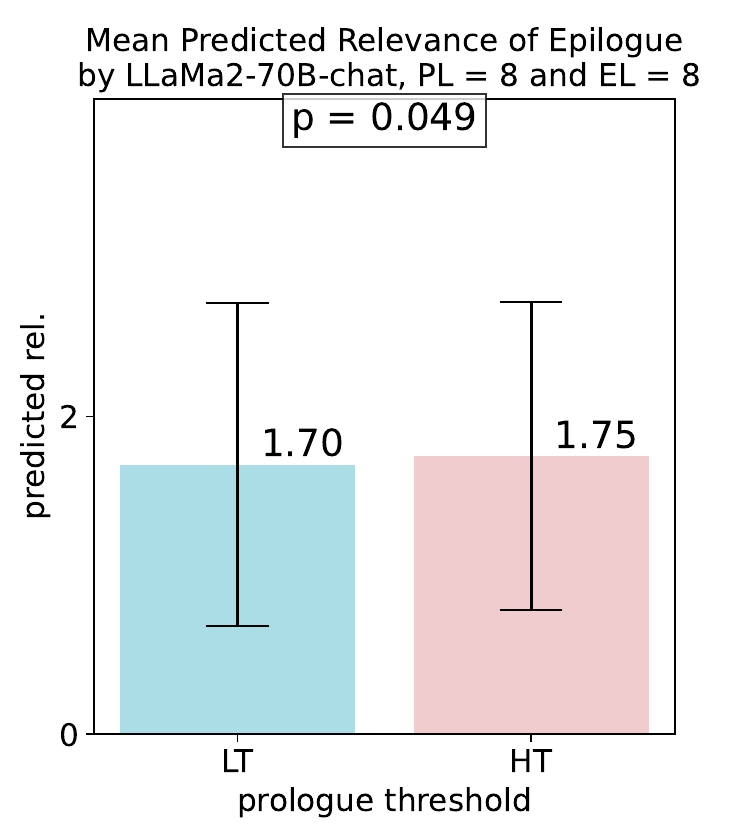}
    \caption*{}
\end{subfigure}
\caption{The average predicted scores for the documents in the epilogue of GPT-3.5, GPT-4o, LLaMa-13B and LLaMa-70B respectively. The results from left to right correspond to GPT-3.5, GPT-4, LLaMa-13B, and LLaMa-70B. From top to bottom, the results are for the conditions with a prologue length (PL) of 4 and an Epilogue length (EL) of 4, a PL of 4 and an EL of 8, and a PL of 8 and an EL of 8. Note that the ground truth relevances of all documents are 2. The p-value is obtained from a dependent t-test.}
\Description{
The average predicted scores for the documents in the epilogue of GPT-3.5, GPT-4o, LLaMa-13B and LLaMa-70B respectively. The results from left to right correspond to GPT-3.5, GPT-4, LLaMa-13B, and LLaMa-70B. From top to bottom, the results are for the conditions with a prologue length (PL) of 4 and an Epilogue length (EL) of 4, a PL of 4 and an EL of 8, and a PL of 8 and an EL of 8. HT represents high threshold, and LT represents low threshold. Note that the ground truth relevances of all documents are 2. The p-value is obtained from a dependent t-test.
}
\label{fig:mean_rel}
\end{figure*}

\begin{table*}[h]
    \centering

\begin{tabular}{cccccccccccccc}
\hline
\multirow{2}{*}{Topic}  & \multirow{2}{*}{Batch Setting} & \multicolumn{3}{c}{GPT-3.5} & \multicolumn{3}{c}{GPT-4o} & \multicolumn{3}{c}{LLaMa2-13B} & \multicolumn{3}{c}{LLaMa2-70B} \\ 
& & HT & LT & p-value & HT & LT & p-value & HT & LT & p-value & HT & LT & p-value \\ \hline
\multirow{3}{*}{168216} 
& PL = 4, EL = 4 & \textbf{1.712} & \textbf{2.412} & <0.001 & \textbf{1.862} & \textbf{2.488} & <0.001 & \textbf{2.000}&\textbf{2.340} & <0.001 & \textbf{1.988} & \textbf{2.425} & <0.001 \\ 
& PL = 4, EL = 8 & \textbf{1.950} & \textbf{2.344} & <0.001 & \textbf{2.066} & \textbf{2.538} & <0.001 &  2.156 & 2.194 & 0.533 & 2.144 & 2.200 & 0.342 \\ 
& PL = 8, EL = 8 & \textbf{1.875} & \textbf{2.450} & <0.001 & \textbf{1.928} & \textbf{2.625} & <0.001 & 2.106 & 2.250 & 0.073 & \underline{2.325} & \underline{1.906} & <0.001\\ \cline{1-14}
\multirow{3}{*}{183378} 
& PL = 4, EL = 4 & \textbf{1.912} & \textbf{2.375} & <0.001 & \textbf{1.681} & \textbf{2.294} & <0.001 &  2.100 & 2.150 & 0.575 & 2.162 & 2.100 & 0.401 \\ 
& PL = 4, EL = 8 & \textbf{1.838} & \textbf{2.125} & <0.001 & \textbf{1.641} & \textbf{2.153} & <0.001 &  2.169 & 2.075 & 0.116 & \underline{2.300} & \underline{1.888} & <0.001\\
& PL = 8, EL = 8 & \textbf{1.738} & \textbf{2.119} & <0.001 & \textbf{1.728} & \textbf{2.372} & <0.001 & 2.275 & 2.225 & 0.485 & \underline{2.106} & \underline{1.794} & <0.001 \\ \cline{1-14}
\multirow{3}{*}{264014} 
& PL = 4, EL = 4 & \textbf{1.012} & \textbf{1.850} & <0.001 & \textbf{0.875} & \textbf{1.738} & <0.001 & \textbf{1.400} & \textbf{2.000} & <0.001  & \textbf{1.412} & \textbf{1.925} & <0.001 \\ 
& PL = 4, EL = 8 & \textbf{1.038} & \textbf{1.912} & <0.001 & \textbf{1.034} & \textbf{1.919} & <0.001 & \textbf{1.781} & \textbf{2.019} & <0.001 & \textbf{1.550} & \textbf{1.850} & <0.001 \\ 
& PL = 8, EL = 8 & \textbf{0.612} & \textbf{1.844} & <0.001 & \textbf{1.009} & \textbf{2.109} & <0.001 & \textbf{1.756} & \textbf{1.975} & <0.002 & 1.656 & 1.719 & 0.453\\ \cline{1-14}
\multirow{3}{*}{443396} 
& PL = 4, EL = 4 & \textbf{0.900} & \textbf{1.225} & 0.003 & \textbf{0.156} & \textbf{0.538} & <0.001 & \textbf{1.550} & \textbf{2.038} & <0.001 & \textbf{1.462} & \textbf{1.862} & 0.002 \\ 
& PL = 4, EL = 8 & \textbf{ 1.062} & \textbf{1.312}& <0.001 & \textbf{0.253} & \textbf{0.550} & <0.001 & \textbf{1.738} & \textbf{2.162} & <0.001 & \textbf{1.644} & \textbf{1.856} & 0.037 \\ 
& PL = 8, EL = 8 & \textbf{0.894} & \textbf{1.494} & <0.001 & \textbf{0.309} & \textbf{1.019} & <0.001 & 1.969 & 2.056 & 0.217 & 1.331 & 1.269 & 0.504 \\ \cline{1-14}
\multirow{3}{*}{451602} 
& PL = 4, EL = 4 & 1.838 & 1.800 & 0.642 & \underline{1.369} & \underline{1.206} & 0.045 & 1.725 & 1.975 & 0.051 & 1.962 & 1.962 & 1.000 \\ 
& PL = 4, EL = 8 & 1.844& 1.856 & 0.817 & \underline{1.509} & \underline{1.272} & <0.001 &  1.906 & 1.956 & 0.515 & 2.088 & 1.938 & 0.053 \\ 
& PL = 8, EL = 8 & 1.812 & 1.838 & 0.699 & \underline{1.797} & \underline{1.966} & 0.015 & \textbf{1.838} &\textbf{2.044} & 0.023 & \textbf{1.750} & \textbf{2.144} & <0.001\\ \cline{1-14}
\multirow{3}{*}{833860} 
& PL = 4, EL = 4 & \textbf{1.338} & \textbf{2.062} & <0.001 & \textbf{1.144} & \textbf{1.956} & <0.001 & \textbf{1.650} & \textbf{2.000} & 0.001 & \textbf{1.325} & \textbf{2.000} & <0.001 \\ 
& PL = 4, EL = 8 & \textbf{1.275} & \textbf{1.819} & <0.001 & \textbf{1.303} & \textbf{2.009} & <0.001 & 1.819 & 1.944 & 0.075 & 1.438 & 1.500 & 0.328 \\ 
& PL = 8, EL = 8 & \textbf{1.200} & \textbf{2.000} & <0.001 & \textbf{1.372} & \textbf{2.209} & <0.001 & \textbf{1.744} & \textbf{1.938} & 0.02 & \underline{1.444} & \underline{1.256} & 0.018\\ \cline{1-14}
\multirow{3}{*}{915593} 
& PL = 4, EL = 4 & \textbf{1.325} & \textbf{1.800} & <0.001 & \textbf{1.269} & \textbf{2.025} & <0.001 & \textbf{1.862} & \textbf{2.100} & 0.014 & \textbf{1.850} & \textbf{2.175} & <0.001 \\ 
& PL = 4, EL = 8 & \textbf{1.400} & \textbf{1.762} & <0.001 & \textbf{1.631} & \textbf{2.241} & <0.001 & \textbf{1.988} & \textbf{2.156} & 0.016 & 1.988 & 2.006 & 0.762 \\ 
& PL = 8, EL = 8 & \textbf{1.356} & \textbf{1.706} & <0.001 & \textbf{1.519} & \textbf{2.519} & <0.001 & 2.031 & 2.094 & 0.439 & 2.019 & 1.962 & 0.552  \\ \cline{1-14}
\multirow{3}{*}{1112341} 
& PL = 4, EL = 4 & 0.738 & 0.738 & 1.000 & \textbf{0.488} & \textbf{0.631} & <0.001 & \textbf{1.412} & \textbf{1.738} & 0.007 & \textbf{0.912} & \textbf{1.350} & <0.001 \\ 
& PL = 4, EL = 8 & 0.831 & 0.856 & 0.629 & \textbf{0.725} & \textbf{0.859} & <0.001 & \textbf{1.556} & \textbf{1.812} & 0.001 & 0.800 & 0.862 & 0.508 \\ 
& PL = 8, EL = 8 & \textbf{0.812} & \textbf{1.112} & <0.001 & \textbf{0.909} & \textbf{1.216} & <0.001 & \textbf{1.762} & \textbf{1.981} & 0.010 & \textbf{0.506} & \textbf{0.938} & <0.001  \\ \cline{1-14}
\multirow{3}{*}{1114819} 
& PL = 4, EL = 4 & \textbf{1.612} & \textbf{2.250} & <0.001 & \textbf{1.025} &

 \textbf{2.088} & <0.001 & \textbf{1.362} & \textbf{2.112} & <0.001 & \textbf{1.662} & \textbf{2.350} & <0.001 \\ 
& PL = 4, EL = 8 & \textbf{1.544} & \textbf{2.100} & <0.001 & \textbf{1.156} & \textbf{2.031} & <0.001 & \textbf{1.794} & \textbf{2.062} & <0.001 & 1.875 & 1.912 & 0.637 \\ 
& PL = 8, EL = 8 & \textbf{1.381} & \textbf{2.081} & <0.001 & \textbf{1.191} & \textbf{2.369} & <0.001 & 1.981 & 2.106 & 0.086 & \underline{1.981} & \underline{1.825} & 0.040 \\ \cline{1-14}
\multirow{3}{*}{1117099} 
& PL = 4, EL = 4 & \textbf{1.275} & \textbf{2.075} & <0.001 & \textbf{0.912} & \textbf{1.712} & <0.001 & \textbf{1.762} & \textbf{2.238} & <0.001  & \textbf{1.800} & \textbf{2.262} & <0.001 \\ 
& PL = 4, EL = 8 & \textbf{1.444} & \textbf{1.944} & <0.001 & \textbf{1.081} & \textbf{1.812} & <0.001 & \textbf{2.144} & \textbf{2.394} & <0.001  & 2.294 & 2.294 & 1.000 \\ 
& PL = 8, EL = 8 & \textbf{1.312} & \textbf{1.994} & <0.001 & \textbf{1.181} & \textbf{2.203} & <0.001 & 2.219 & 2.325 & 0.173 &\underline{2.400} & \underline{2.169} & 0.003  \\ \cline{1-14}
\end{tabular}

    \caption{Results by topic, from left to right correspond to GPT-3.5, GPT-4o, LLaMa-13B, and LLaMa-70B. From top to bottom, the results are for the conditions with a prologue length (PL) of 4 and an epilogue length (EL) of 4, a PL of 4 and an EL of 8, and a PL of 8 and an EL of 8. The p-value is obtained from a dependent t-test. HT represents high threshold, and LT represents low threshold. Bold indicates that document relevance in epilogues is significantly higher under LT treatment compared to HT (p < 0.05); underline indicates that document relevance in epilogues is significantly higher under HT treatment compared to LT (p < 0.05).}
    \label{tab:results}
\end{table*}

Figure~\ref{fig:mean_rel} shows the scores assigned to the documents in the epilogue by different models under different batch setups when the prologue is either low threshold (LT) or high threshold (HT). For each combination, we calculated the average relevance scores of two groups of epilogues under each theme, one processed by LT and the other by HT. We then conducted a dependent t-test (please note that both groups of epilogues were composed of the same documents in the same order).

From Figure~\ref{fig:mean_rel} one can observe that regardless of the batch setup, the scores from GPT-3.5, GPT-4, and LLaMa2-13B are generally influenced by threshold priming. Specifically, when the relevance threshold of the documents in the prologue is higher, these models tend to assign lower scores to the documents in the epilogue, and this difference is statistically significant (p < 0.001). This trend is consistent with the observations made by Scholer~\textit{et al.}~\citep{scholer2013}. Additionally, one can also observe that LLaMa2-13B exhibits a slighter difference in relevance assessment of documents in the epilogue under LT and HT compared to the GPT series. 

However, for LLaMa2-70B, this trend can only be observed when PL = 4 and EL = 4. When PL = 4 and EL = 8, LLaMa2-70B does not show a significant difference in relevance judgment of documents in the epilogue under LT and HT (p = 0.434). When PL = 8 and EL = 8, LLaMa2-70B tends to assign higher relevance scores to documents in the epilogue under HT (p = 0.049), which is contrary to the trend observed by Scholer \textit{et al.}~\citep{scholer2013}. 

Regarding \textbf{RQ1}, we found that, when the prologue is 4 and the epilogue is 4, GPT-3.5, GPT-4o, LLaMa2-13B, and LLaMa2-70B are all affected by threshold priming. Moreover, GPT-3.5 and GPT-4o are more significantly influenced by this effect compared to LLaMa2-13B and LLaMa2-70B.

Table~\ref{tab:results} further presents the assessment results of each model on different topics under various batch setups. From Table~\ref{tab:results}, one can observe that:

\begin{itemize}
    \item When PL = 4 and EL = 4, GPT-3.5 tends to assign higher relevance scores to the documents in the epilogue under LT in 8 out of 10 topics (p < 0.05); GPT-4 tends to assign higher relevance scores under LT in 9 topics (p < 0.05); LLaMa2-13B tends to assign higher relevance scores under LT in 8 topics (p < 0.05); LLaMa2-70B tends to assign higher relevance scores under LT in 8 topics (p < 0.05). Notably, in topic 451602, none of the four models exhibited a tendency to assign higher relevance scores to the documents in the epilogue under LT. In fact, GPT-4 even tended to assign lower scores (p = 0.045).
    \item When PL = 4 and EL = 8, GPT-3.5 tends to assign higher relevance scores to the documents in the epilogue under LT in 8  topics (p < 0.05); GPT-4 tends to assign higher relevance scores under LT in 9 topics (p < 0.05); LLaMa2-13B tends to assign higher relevance scores under LT in 6 topics (p < 0.05); LLaMa2-70B showed no statistically significant difference in the average relevance scores assigned to the documents in the epilogue under LT and HT in 7 topics. In topic 451602, GPT-4 still tends to assign lower relevance scores under LT (p < 0.001). Similarly, LLaMa2-70B also assigns slightly lower relevance scores under LT, but the difference is not significant (p = 0.328).  
    \item When PL = 8 and EL = 8, GPT-3.5 tends to assign higher relevance scores to the documents in the epilogue under LT in 9 topics (p < 0.05); GPT-4 tends to assign higher relevance scores under LT in 9 topics (p < 0.05); LLaMa2-13B tends to assign higher relevance scores under LT in 4 topics (p < 0.05), and  in 5 out of the 6 remaining topics (excluding topic 188378), LLaMa2-70B assigns slightly higher relevance scores under LT compared to HT, but the differences are not significant.; In contrast to these models, LLaMa2-70B tends to assign higher relevance scores to the documents in the epilogue under HT in 5 topics (p < 0.05). 
\end{itemize}

These results indicate that the \textit{degree} and the \text{direction} of threshold priming effect in large language models can be influenced by different \textit{batch setups} when performing batch relevance judgments on documents. For LLaMa2-70B under PL = 4 and EL = 4, our observed results are contrary to those reported by Scholer~\textit{et al.}~\citep{scholer2013}. One possible explanation is that this is similar to the anchoring effect observed by Shokouhi~\textit{et al.}~\citep{shokouhi-2015} in relevance assessment tasks. Shokouhi~\textit{et al.}~\citep{shokouhi-2015} noted that if the first document has a high relevance score, assessors tend to assign higher relevance scores to the second document. However, further evaluation experiment is required to empirically validate this hypothesis.

Regarding \textbf{RQ2}, we found that the extent to which LLMs are influenced by threshold priming varies across different topics. In some topics (\textit{e.g.,} 451602), most models did not exhibit significant differences in relevance scores under different threshold priming conditions, or they exhibited results more akin to anchoring effect, which needs further exploration in future research across different topics, bias conditions, user populations and search modalities.  
\section{Discussion}

This work, inspired by insights from behavioral economics, investigated the threshold priming bias in LLMs within the context of information retrieval system evaluation. The findings of this work contribute to a deeper understanding of AI systems' judgments and decision-making processes from an interdisciplinary perspective, and offer practical implications for designing effective bias mitigation strategies and human-centered unbiased AI systems. 

Regarding our RQs, we have the following findings:
\begin{itemize}
    \item Regarding \textbf{RQ1}, we found that, when the prologue is 4 and the epilogue is 4, GPT-3.5, GPT-4o, LLaMa2-13B, and LLaMa2-70B are all affected by threshold priming. Moreover, GPT-3.5 and GPT-4o are more significantly influenced by this effect compared to LLaMa2-13B and LLaMa2-70B.
    \item Regarding \textbf{RQ2}, we found that the extent to which LLMs are influenced by threshold priming varies across different topics. In some topics (\textit{e.g.,} 451602), most models did not exhibit significant differences in relevance scores under different threshold priming conditions, or they exhibited results more akin to the anchoring effect, which needs further exploration in future research.
\end{itemize}

While our study has contributed valuable insights into the threshold priming effect in LLMs, it also presents certain limitations. One limitation is the restricted variety of topics and document batches tested, which may not fully capture the diversity of real-world applications. Moreover, the reliance on specific LLM models could skew the generalizability of our findings. However, these limitations do not substantially detract from the value of our experiments. To mitigate these issues, we have employed robust statistical methods to ensure the reliability of our findings and plan to extend our research to more diverse datasets, threshold priming and reference conditions, different prompt structures for assessment tasks, and additional LLM configurations in future studies. This approach will help us explore the broader applicability of our results, further refine our understanding of LLM behavior in complex decision-making scenarios, and better connect the knowledge about cognitive biases with AI bias mitigation techniques~\cite{azzopardi2024search, liu2024search}.

\section{Conclusion}
This study paves the path toward further explorations on threshold priming as well as other related biases and open questions of LLMs in IR evaluation tasks. For instance, what are the boundaries for triggering threshold priming? How are LLM's relevance and credibility labels influenced by threshold priming affect system rankings in offline evaluations? Additionally, as observed in our experiments, LLMs may be subject to cognitive biases beyond threshold priming, such as the anchoring bias, decoy effect, and reference dependence, during relevance assessment tasks. Future research directions include identifying and quantifying the biases (including the cognitively biased judgments) occurring in LLM's decisions to better understand their limitations, and exploring methods to reduce these biases and facilitate more reliable human-AI collaboration across varying tasks and problem spaces, user cognitive states, and interaction modalities~\cite{markwald2023constructing, liu2020identifying, kiesel2021meant}.

Similar to searchers with cognitive limits~\cite{liu2023behavioral}, LLMs, contrary to the widely-accepted rationality assumptions, may also be "boundedly rational" to some extent. Based on the results observed in our experiments, we suggest that the IR community have the LLM judge the same set of documents in different topics, random orders (\textit{i.e.,} multiple trials) as well as relevance levels from each trial. Additionally, future research can explore how prompting engineering and fine-tuning techniques can help identify and effectively mitigate the cognitive biases of LLMs in automated judgment tasks, and how we can leverage the expertise of (instead of replacing) domain experts and information professionals in supervising and debiasing the judgments of LLM assessors in varying scenarios. 

\balance
\bibliographystyle{ACM-Reference-Format}
\bibliography{myref}

\end{document}